\documentclass{article}

\pdfoutput=1

\PassOptionsToPackage{numbers, compress}{natbib}

\usepackage[final]{nips_2018}

\usepackage[para]{footmisc}

\usepackage[utf8]{inputenc} 
\usepackage[T1]{fontenc}    

\usepackage{hyperref}
\usepackage{url}

\usepackage{booktabs}       
\usepackage{amsfonts}       
\usepackage{nicefrac}       
\usepackage{microtype}      
\usepackage{subfigure} 
\usepackage{graphicx} 

\usepackage{bm}
\usepackage{amsmath}
\usepackage{amsthm}
\usepackage{pdfpages}

\usepackage{xcolor}

\title{ML for Flood Forecasting at Scale}

\author{
  Sella Nevo ^ {\ref{gr}}\ \  \textbf{Vova Anisimov} ^ {\ref{gr}}\ \   \textbf{Gal Elidan} ^ {\ref{gr_huji}}\ \  \textbf{Ran El-Yaniv} ^ {\ref{gr_tech}}\ \ \textbf{Pete Giencke} ^ {\ref{gcr}}\ \  \\
  \textbf{Yotam Gigi} ^ {\ref{gr_huji}}\ \ \textbf{Avinatan Hassidim} ^ {\ref{gr_bi}}\ \ \textbf{Zach Moshe} ^ {\ref{gr}}\ \ \textbf{Mor Schlesinger} ^ {\ref{gcr}}\ \ \textbf{Guy Shalev} ^ {\ref{gr}}\ \ \\
  \textbf{Ajai Tirumali} ^ {\ref{gr}}\ \ \textbf{Ami Wiesel} ^ {\ref{gr_huji}}\ \ \textbf{Oleg Zlydenko} ^ {\ref{gcr}}\ \  \textbf{Yossi Matias} ^ {\ref{gr},\ref{gcr}}
}

\begin{document}

\maketitle
\begin{abstract}
    Effective riverine flood forecasting at scale is hindered by a multitude of factors, most notably the need to rely on human calibration in current methodology, the limited amount of data for a specific location, and the computational difficulty of building continent/global level models that are sufficiently accurate. Machine learning (ML) is primed to be useful in this scenario: learned models often surpass human experts in complex high-dimensional scenarios, and the framework of transfer or multitask learning is an appealing solution for leveraging local signals to achieve improved global performance. We propose to build on these strengths and develop ML systems for timely and accurate riverine flood prediction. 
    \footnotetext[1]{\label{gr} Google Research} 
    \footnotetext[2]{\label{gr_huji} Google Research and The Hebrew University}
    \footnotetext[3]{\label{gr_tech} Google Research and Technion - Israel Institute of Technology}
    \footnotetext[4]{\label{gr_bi} Google Research and Bar-Ilan University}
    \footnotetext[5]{\label{gcr} Google Crisis Response}
\end{abstract}

Floods are the most common and deadly natural disaster in the world. Every year, floods cause from thousands to tens of thousands of fatalities \cite{cred, jonkman2003loss, unisdr, jonkman2005global, doocy2013human}, affect hundreds of millions of people \cite{doocy2013human, jonkman2005global, unisdr}, and cause tens of billions of dollars worth of damages \cite{cred, unisdr}. These numbers have only been increasing in recent decades \cite{loster1999flood}. Indeed, the UN charter notes floods to be one of the key motivators for formulating the sustainable development goals (SDGs), and directly challenges us: "They knew that earthquakes and floods were inevitable, but that the high death tolls were not." \cite{undp}

Early warning systems for floods, even with limited lead time and accuracy, can reduce both fatalities and economic damages by over a third, and in some cases almost by half \cite{who, pilon1998guidelines, worldbank}. Unfortunately, the majority of human costs that are due to flooding are concentrated in developing countries \cite{doocy2013human}, which often lack effective and actionable early warning systems due to limited data collection, funding, or professional expertise \cite{stromberg2007natural}. The result is that, across multiple countries, thousands die on average every year, and relief and mitigation efforts have little information to rely on.

We focus on riverine floods which are responsible for much of the effect on human life, and for which we believe the opportunity for ML-based predictive models is substantial. Perhaps the biggest challenge in scaling current hydrologic modeling efforts to a global scale is the parameter calibration process - an optimization process aimed at matching model outputs to ground-truth measurements. The standard existing methodology involves significant manual work, which is arduous and time consuming \cite{boyle2000toward, eckhardt2001automatic}, difficult to teach \cite{boyle2000toward}, and is often over-parameterized \cite{box2015time}. The result is that models that appear to work well at construction time, are often not able to generalize because of the complex and dynamic nature of the problem \cite{beven1989changing}. Indeed, there have been many attempts to automate this process, but without being able to reach a level of accuracy and reliability that is sufficient for use in operational systems \cite{boyle2000toward, madsen2002comparison}. Practical use has thus been limited to either semi-automatic processes which include manual intervention or interaction, \cite{boyle2000toward, gupta1999status}, or to aggressive constraining of the model that leads to inferior results \cite{beven2011rainfall, eckhardt2001automatic}.

The field of Machine Learning (ML) is incredibly well-placed to overcome the above limitations since many of the core challenges that stand in the way of automatic calibration of hydrologic models are central to the ML arsenal. In fact, most large-scale ML systems these days replaced and improved on earlier inferior system that were built and calibrated by hand. There have been previous efforts at using ML for flood prediction, but these have been typically limited to a handful of basins or even a single one \cite{shrestha2006machine, tokar1999rainfall, dawson2001hydrological, hsu2002self, campolo1999river, jain2007hybrid, hsu1995artificial, chen2015comparative}. More often than not, the learned model was not able to compete with manual calibration or achieve sufficient reliability for operational use. We believe that substantially better models can be learned by utilizing data at a larger scale. We also believe that recent advances in inductive transfer, transfer learning and multitask learning (see \cite{Pan:2010} for a recent survey) can overcome the difficulty of learning effective models from multiple related sources.

One of the main reasons the potential of ML for flood prediction has not yet been realized is that there are few organizations (or even communities) that have all the necessary components for success under one roof: knowledge and experience in operational flood forecasting; the ability to centralize data at a global scale; the required expertise in ML; the ability to effectively deploy predictive information.

To start changing this reality, we  developed and recently deployed an operational hydro-dynamic model, in a pilot area around Patna, India \cite{blogpost}, with the explicit goal of preparing the ground for integrating ML models into the process. In this model, the physical processes are relatively well understood for several decades now \cite{wu2007computational}, and relatively little calibration is required. It receives real-time measurements and short-term forecasts of river water levels in real-time, and produces an inundation map estimating the flood extent. Based on past events, our model’s flood extent estimates are accurate down to a spatial resolution of 300 meters, achieving over 90\% recall and over 75\% precision. These results have been borne out by our first alerts produced in real time for the 2018 monsoon season. While our initial results are promising for this "simple" component of flood prediction, we did run into challenges we believe machine learning is well placed to solve: high computational costs of the physics-based simulations; inaccuracies due to error-prone and biased inputs; manual calibration that still dominates the amount of work required for each new location.

The above launched hydro-dynamic model is just one component of a full flood forecasting system, and the best understood one at that. We believe ML can improve the quality of multiple components, including ones that are not well simulated by physics-based models, from absorption and snow-melt models to river discharge estimation. Toward making this possible, we are collecting, organizing  and combining open data sets from different sources to make this problem more accessible for the ML community. Together with the launched hydro-dynamic model, we believe this sets the ground for the development, assessment, and deployment of ML models at the scale of countries and continents.

Concretely, we believe ML can provide large contributions to operational flood forecasting systems along each of the following axes:
\begin{description}
 \item[ML-based hydro-dynamic modeling]  Computational costs still limit the scope and the resolution of hydro-dynamic models. These costs could potentially be reduced by up to orders of magnitude by integrating ML-based derivation methods, as has been experimentally done for other fluid dynamics systems (see e.g. \cite{bar2018data}).
 
 \item[Remote discharge estimation] Limited data is perhaps the most substantial obstacle for incorporating ML into the hydrology field. Gauges measuring discharge and water levels are relatively scarce, on the order of 100,000 globally. Yet the quantity, quality and variety of satellites constantly imaging all rivers is rising at a rapid pace. Improving our ability to estimate river discharge \emph{without} in-situ measurements is a task that is both critical for the hydrology field, and incredibly well suited for ML. See \cite{yotam2018towards} for our preliminary work in this area. 
 
 \item[ML hydrologic models] Data-driven hydrology models often under-perform, mainly due to lack of sufficient data: for each basin there are typically only hundreds to thousands of measurements. Yet, hydrologic phenomena follows the same physical principles everywhere, and it is reasonable to assume that by jointly considering the tasks of flood prediction at multiple locations we can do better. We conjecture that by utilizing recent advances in transfer learning, we can do just that, and effectively learn at a global scale.
 
 \item[Hybrid physics-ML hydrologic models] An alternative to the pure ML approach above, is to adopt a hybrid approach where an ML model is integrated with classical physics-based components, which can contain important prior knowledge about the structure of the domain. Intuitively, one could view this approach as allowing the conceptual components to capture the majority of the hydrologic modeling, while the ML model is responsible for calibration, error corrections, and perhaps additional processes that were not well modelled.

\end{description}
In summary, we believe ML is particularly suited to improve effective flood prediction at scale, and we are excited to catalyze more academic and non-academic efforts in this direction.



\bibliography{bib_aistats}
\bibliographystyle{plain}

\end{document}